\documentclass{article}

\usepackage[preprint]{nips_2018}

\usepackage[utf8]{inputenc} 
\usepackage[T1]{fontenc}    
\usepackage{hyperref}       
\usepackage{url}            
\usepackage{booktabs}       
\usepackage{amsmath,amsthm,amssymb,amsfonts}       
\usepackage{nicefrac}       
\usepackage{microtype}      
\usepackage{graphicx}
\usepackage{color}

\renewcommand{\vec}[1]{\mathbf{#1}}

\usepackage{algpseudocode}
\usepackage{algorithm}
\usepackage{bbm}
\usepackage{verbatim}

\newcommand{\captionsize}{\footnotesize}

\title{Complex Relations in a Deep Structured Prediction Model for Fine Image Segmentation}
\author{
  Cristina Mata, ~Guy Ben-Yosef, ~Boris Katz \\
 Computer Science and Artificial Intelligence Laboratory\\
 Center for Brains, Minds and Machines\\
 MIT\\
 \texttt{\{cfmata, gby, boris\}@csail.mit.edu} \\
 }

\begin{document}

\maketitle
\begin{abstract}
Many deep learning architectures for semantic segmentation involve a Fully Convolutional Neural Network (FCN) followed by a Conditional Random Field (CRF) to carry out inference over an image. These models typically involve unary potentials based on local appearance features computed by FCNs, and binary potentials based on the displacement between pixels. We show that while current methods succeed in segmenting whole objects, they perform poorly in situations involving a large number of object parts. We therefore suggest incorporating into the inference algorithm additional higher-order potentials inspired by the way humans identify and localize parts. We incorporate two relations that were shown to be useful to human object identification – containment and attachment – into the energy term of the CRF and evaluate their performance on the Pascal VOC Parts dataset. Our experimental results show that the segmentation of fine parts is positively affected by the addition of these two relations, and that the segmentation of fine parts can be further influenced by complex structural features.
\end{abstract}


\section{Introduction}
\label{sec:introduction}
Semantic segmentation is an important task in computer vision that comprises a subset of structured prediction methods. In recent years it has garnered much attention due to its use in street scene understanding for autonomous vehicles, medical image analysis, and a host of other applications for computer vision. The use of deep convolutional neural networks has resulted in substantial improvements for object-level segmentation. When used in conjunction with undirected graphical models, most commonly conditional random fields (CRFs), performance increases \cite{Chen_etal_2018_TPAMI}. However, an overview of these results shows that there is considerable room for improvement, particularly for segmentation at the parts-level.
The detailed, or fine segmentation of parts at the richness and accuracy of human-level can be crucial for visual tasks involving the understanding of actions and interactions \cite{Ben-Yosef_etal_2018_Cognition,Ben-Yosef_Ullman_2018_PRSIF}.

Fine part segmentation remains a hard problem because most objects contain a large number of parts that may be arranged in many different configurations, and small parts often appear with low visibility. One fundamental issue with most current part segmentation methods is that boundary localization is poor. This results in segmentations filled with errors a human would never make, such as segmenting a horse's torso into disjoint parts or making its head disproportionately large. While CRFs based on unary and binary potentials were introduced with boundary refinement in mind, they do not capture the complex relations that humans use to identify object parts. 

We propose incorporating potentials that express these relations directly into the structure of the CRF, specifically, \textit{containment} and \textit{attachment}, two complex relations that were shown to be useful to human identification of object parts \cite{Ben-Yosef_etal_2018_Cognition}. The main challenge in adding these higher-order potentials to the CRF is carrying out inference over the altered structure. We use mean field variational inference to determine the most likely label for each pixel in an image, and derive update equations due to our potentials. 
An outline of our paper follows: in section 2 we discuss related work in the field of semantic segmentation. In section 3 we outline our complex potentials and present their update equations. In section 4 we present our results on the Pascal-Parts dataset \cite{Chen_etal_2014_CVPR} and conclude in section 5 with a discussion.

\section{Related work}
\label{sec:related_work}
Significant progress has been made on semantic segmentation through the introduction of end-to-end Convolutional Neural Networks e.g.,  \cite{Long_etal_2015_CVPR,Shelhamer_etal_2017_TPAMI,Chen_etal_2018_TPAMI}, by taking a network architecture that was originally designed for the ImageNet object classification task and adapting it to a `Fully Convolutional Network' (FCN) in which the output map is equal in size and shape to the input layer. Similar to classification CNNs, FCNs incorporate  pooling and convolutional layers, which cause resolution of the original visual signal to be gradually decreased with the depth of the network  during the encoding stage. Then, up-sampling techniques and skip connections are used to recover the resolution of the input map  during a decoding stage. As a result, the accuracy of boundary localization is limited, particularly when the segmented object or part appears with low resolution in the input image.

Much of the recent research on semantic segmentation has focused on the decoding stage: increasing the boundary accuracy while also providing a correct and coherent labeling of the objects or parts. The results of this research include sophisticated decoding architectures for encoder-decoder segmentation networks. For example, SegNet \cite{Badrinarayanan_etal_2017_TPAMI} stores indices of max-pooling layers in the encoding stage, and uses them later for ``unpooling" layers in decoding. UNet \cite{Ronneberger_etal_2015_ICMI} concatenates feature maps in encoding stage to upsampled feature maps in decoding stage, and RefineNet \cite{Lin_etal_2017_CVPR} which employs long-range residual connections for identity mapping, to pass high-resolution features from earlier convolutions to deeper layers. Even so, progress on semantic segmentation of objects and parts is slow; one reason for this might be that current systems do not take into account what is known about human identification and interpretation of object and parts.

A successful approach to address resolution limitations in semantic segmentation is to combine FCNs with an additional graphical model inference stage, most commonly a fully connected CRF in which each image pixel is a node, and every pair of nodes is connected, termed a dense CRF \cite{Krahenbuhl_Koltun_2011_NIPS}. The energy equation for the CRF is typically based on the Gibbs energy of an Ising model and contains unary and binary potentials.

While the most successful approach for semantic segmentation to date is based on an FCN + CRF with powerful unary potentials provided by the FCN \cite{Chen_etal_2018_TPAMI}, recent work has shown that adding potentials that depend on more than two pixels, termed higher-order potentials, is useful for the general segmentation task (e.g., \cite{Kohlie_etal_2007_CVPR,Lin_etal_2016_CVPR,Arnab_etal_2016_ECCV}). 
\cite{Kohlie_etal_2007_CVPR} extend the class of models on which efficient
approximate inference may be performed over CRFs to the $P^{n}$ Potts model, and \cite{Arnab_etal_2016_ECCV} build off of this pattern-based potential formulation by including a coherence term based on superpixels as a layer within the FCN rather than within the dense CRF inference module, and show improved performance for semantic segmentation. 

While the terms in the extended energy equation combine intuitive notions of coherency between local regions of an image, they do not encompass more complex semantic relations between object parts that humans use to identify objects. Motivated by the role of complex relations in cognitive models for object image interpretation \cite{Ben-Yosef_etal_2018_Cognition}, and by the success of recent semantic segmentation algorithms to leverage performance via additional higher order potentials, we turn to study new complex relation terms in an FCN followed by the dense CRF framework for semantic part segmentation. As will be shown below, the contribution of such relation terms, which are inspired by cognitive models, is significant to part segmentation performance, particularly when the number of parts is high.

\section{Adding more complex relations} 

We are focusing here on two relation terms, the {\em containment} of parts and the {\em attachment} of parts, which were shown to be useful for replicating human object interpretation \cite{Ben-Yosef_etal_2018_Cognition}, and a derivation of their mean field update rule. We then use the derived rules for implementing a dense CRF model (Sec. \ref{sec:experimental_results}) that includes the basic unary and binary potentials, with the addition of the novel, more complex potential terms derived here. Note that higher-order potential terms, beyond the ones discussed here, could be also considered for improving semantic part segmentation.

\par We begin by introducing notation to describe the dense CRF and our potentials. Our goal is to find an optimal labeling for the set of latent variables $X = \{ x_{1}, x_{2}, ..., x_{N} \}$ where each $x_{i}$ is a node in our CRF representing a pixel in an image. Each variable takes on a label in the set $\mathcal{L} = \{ 1, 2, 3, ..., L\}$, and $P(X)$ denotes the true distribution of label assignments to pixels for a single image. 
\par Exact inference on $X$ is intractable, so we use mean field variational inference to find a distribution $Q(X)$ that is a good approximation to the true, or target distribution, $P(X)$. In the following derivation we assume that $Q(X)$ satisfies the mean field approximation, i.e., $Q(X) = \prod \limits_{i=1}^{N} Q_{i}(x_{i})$ where $Q_{i}(x_{i})$ is the distribution over $\mathcal{L}$ for pixel $x_{i}$ with $\sum \limits_{l=1}^{L} Q_{i}(x_{i} = l) = 1$.

\par The energy of the basic dense CRF is

\begin{equation}
\label{eq:basic_dense_crf}
E(x) = \sum \limits_{i} \psi_{u}(x_{i}) + \sum \limits_{i, j} \psi_{p}(x_{i}, x_{j})
\end{equation}

where the first term is the unary potential and is measured over each pixel, and the second term is taken over every pair of pixels \cite{Krahenbuhl_Koltun_2011_NIPS}.

Using the KL-divergence to measure the distance between $Q(X)$ and $P(X)$ leads to the following update equation for each distribution $Q_{i}(x_{i})$ that depends on the potentials in the energy \cite{KF}:

\begin{equation}
Q_{i}(x_{i} = l) = \frac{1}{Z_{i}} \exp \left \{ -\sum \limits_{\psi:x_{i} \in U_{\psi}} \sum \limits_{c \in \mathcal{C}} \sum \limits_{\vec{x}_{c} | x_{i} = l} Q_{c-i}(\vec{x}_{c-i}) \psi_{c}(\vec{x}_{c}) \right \}
\label{eq:mean_field_update}
\end{equation}

where the partition function $Z_{i}  = \sum \limits_{l} \exp \left \{ -\sum \limits_{\psi:x_{i} \in U_{\psi}} \sum \limits_{c \in \mathcal{C}} \sum \limits_{\vec{v}_{c} | v_{i} = l} Q_{c-i}(\vec{v}_{c-i}) \right \}$ normalizes the expression. The first sum is taken over all potential functions $\psi$ for which pixel $x_{i}$ is in the domain $U_{\psi}$ of the function. The second sum is taken over all cliques $c \in \mathcal{C}$, the total set of cliques in the CRF, where a clique is a complete subgraph of the CRF. The third sum is taken over $\vec{x}_{c}$, the set of all configurations of pixels in the clique when $x_{i} = l$. In the body of the expression we sum over $Q_{c-i}(\vec{x}_{c-i})$, the joint distribution over the variables $\{x_{j} \in c : j \neq i\}$ when $x_{i} = l$, multiplied by the value $\psi_{c}(\vec{x}_{c})$ of the potential applied to clique $c$.

To increase coherency within local image regions and promote boundary localization, a superpixel potential was added to Eq. \ref{eq:basic_dense_crf} by \cite{Arnab_etal_2016_ECCV}, namely:

\begin{equation}
\psi_{\text{s}}(\vec{x}_{c}) = \begin{cases}
    w_{\text{low}}(l), & \forall i \in c, \vec{x}_{c} = l\\
    w_{\text{high}}, & \text{otherwise}
  \end{cases}
\label{eq:superpixel_potential}
\end{equation}

where each clique $c$ is a set of pixels in a superpixel, and the term enforces that a low weight is added to the energy when all pixels in a clique take on the same label. Our terms build off of this pattern-based potential by changing the condition for adding a low weight. The corresponding update rule for this potential is

\begin{equation}
\sum \limits_{\vec{x}_{c} | x_{i} = l} Q(\vec{x}_{c-i})\psi_{s}(\vec{x}_{c}) =  \sum \limits_{\vec{x}_{c} | x_{i} = l}  \Big(\prod \limits_{j \in c, j \neq i} Q_{j}(x_{j} = l) \Big) *w_{\text{low}}(l) + \Big( 1 - \Big( \prod \limits_{j \in c, j \neq i} Q_{j}(x_{j} = l) \Big) \Big)* w_{\text{high}}
\end{equation}

\subsection{Containment}
\label{sec:containment}

\par We consider the containment potential

\begin{equation}
\psi_{\text{i/o}}(\vec{x}_{c}) = \begin{cases}
    w_{\text{low}}(l), & \forall i \in c, \vec{x}_{c} = \text{OR}(l,l'(c))\\
    w_{\text{high}}, & \text{otherwise}
  \end{cases}
\label{eq:inside_outside_potential}
\end{equation}

in which we assign a low weight $w_{\text{low}}(l)$ to the clique $c$ when, for all variables $x_{i} \in c$, $x_{i} = l$ or $x_{i} = l'$. In all other cases we assign a high weight $w_{\text{high}}$. Note that $w_{\text{low}}$ depends on the label $l$ of the pixel we are updating over, as well as some label $l'$ that we determine prior to inference and which depends on the clique. In our formulation for segmentation, each clique $c$ is the boundary of a superpixel, and $l'$ is the most common label within that superpixel. At runtime the value for $w_{\text{low}}$ may be determined from a look-up table as described below.

\par
\begin{figure}
\centering
\includegraphics[width=.2\linewidth]{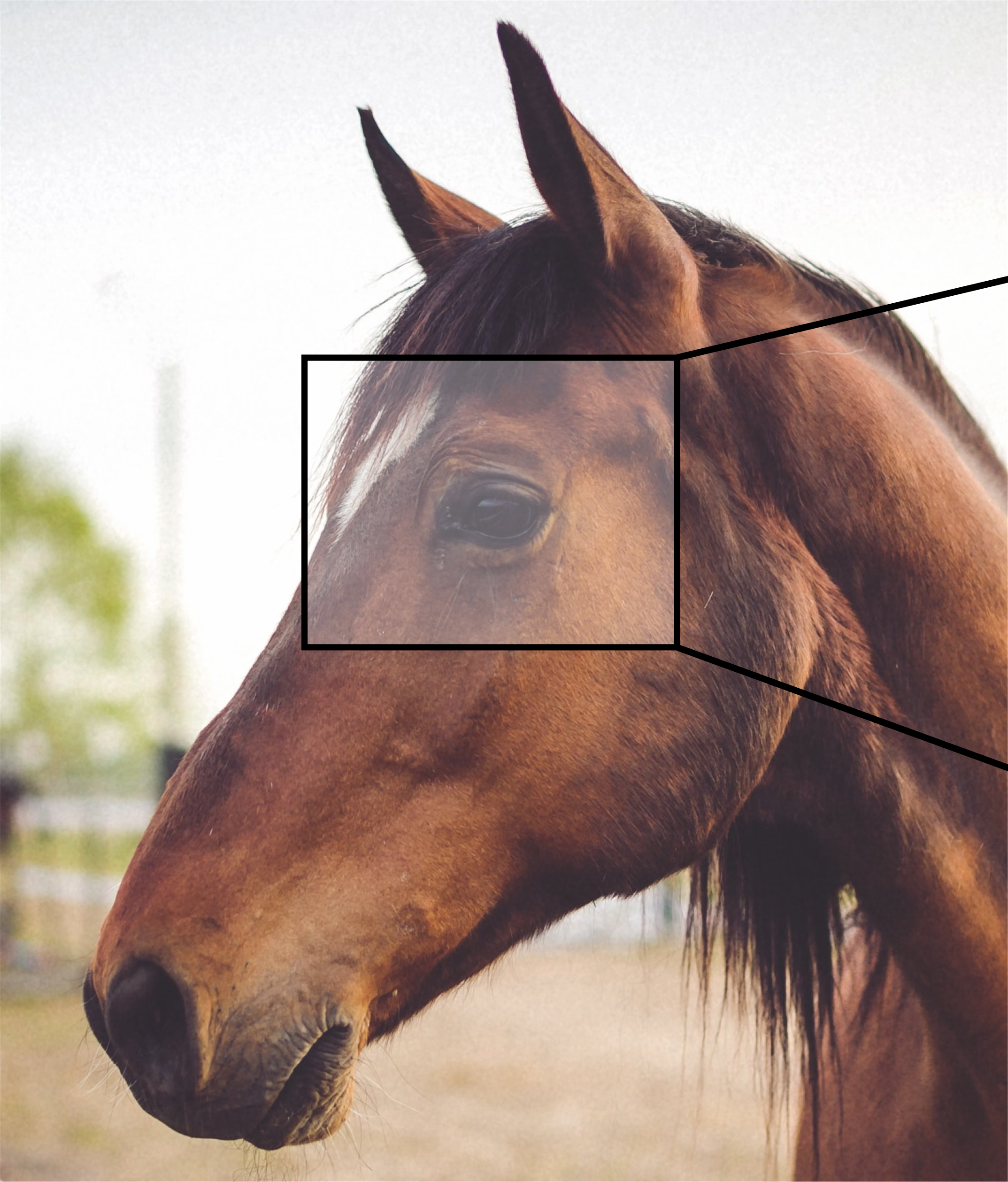}
\includegraphics[width=.35\linewidth]{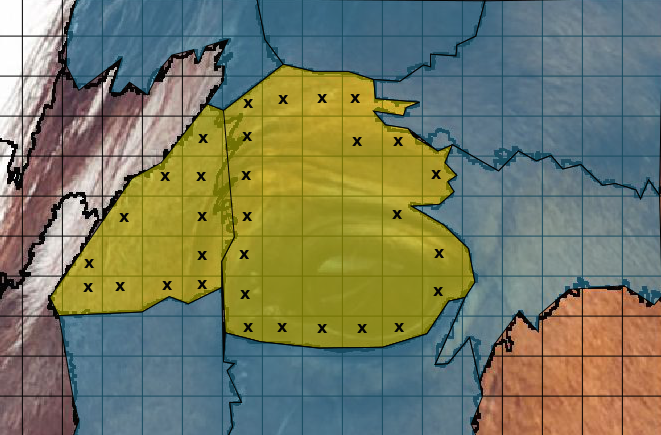}
\includegraphics[width=.35\linewidth]{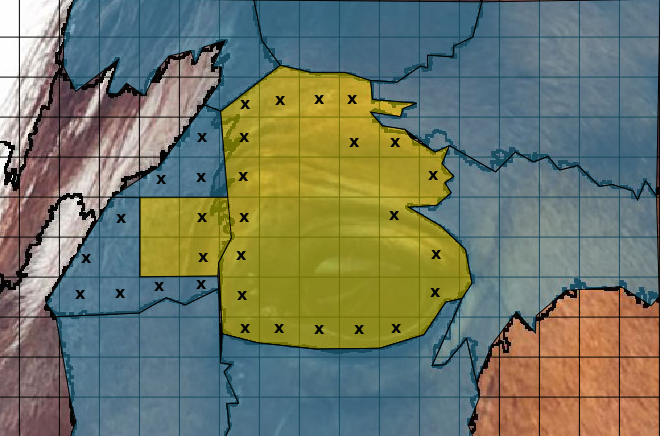}
\caption{\captionsize
       Error correction using the containment potential. Left: The original image. Middle: The blue marks areas labeled head, the yellow is labeled eye, and we see that the eye is over-segmented. Each x marks a pixel along the boundary, which together constitute a clique that containment is calculated over. Right: The corrected labeling.}
\label{fig:containment_correction}
\end{figure}

 As an example, say we are considering the boundary of a superpixel whose most common label $l'$ is ``left eye". As we update the value of each pixel in the boundary we will consider all possible labels $l$ for each pixel. If $l$ is ``left eye" or ``head" then $w_{\text{low}}(l) \approx 0$; however if $l$ is any other label such as ``torso" then $w_{\text{low}}(l) = w_{\text{high}}$, since intuitively we don't want to enforce any containment relation between the eyes and other labels. These relationships between labels are determined beforehand or learned and then stored in a look-up table. Learning a relation involves measuring its presence in the dataset. Once a relation potential is well-defined, it is straightforward to measure its presence between every pair of labels by calculating the potential directly for each image in the ground truth training set. Once a proportion of images containing the relation has been formed for each pair of labels, a threshold may be set, above which it is determined that a relation between the pair of labels exists. Then these relations may be used during inference. Figure \ref{fig:containment_correction} demonstrates exactly which pixels in a superpixel would be included in a containment clique.

\par Plugging $\psi_{\text{i/o}}$ into Eq. \ref{eq:mean_field_update} gives the following update for each clique $c$:

\begin{equation*}
\sum \limits_{\vec{x}_{c} | x_{i} = l} Q_{c-i}(\vec{x}_{c-i}) \psi_{\text{i/o}}(\vec{x}_{c}) = \sum \limits_{\vec{x}_{c} | x_{i} = l} Q_{c-i}(\vec{x}_{c-i}) * w_{\text{low}}(l) + \sum \limits_{\vec{x}_{c} | x_{i} = l} (1 - Q_{c-i}(\vec{x}_{c-i})) * w_{\text{high}} 
\end{equation*}

Consider the first term on the right hand side of the above. This term encompasses situations in which $x_{i} = \text{OR}(l,l')$ for all $x_{i} \in c$, in which case we know the likelihood of this happening can be calculated as 

\begin{equation*}
\prod \limits_{j \in c, j \neq i}Q_{j}(x_{j} = \text{OR}(l,l')) = \prod \limits_{j \in c, j \neq i} Q_{j}(x_{j} = l) + Q_{j}(x_{j} = l')
\end{equation*}

The second term encompasses all other configurations of the pixels in the clique, so this may be calculated as the complement of the above:
$1 - \Big( \prod \limits_{j \in c, j \neq i} Q_{j}(x_{j} = l) + Q_{j}(x_{j} = l') \Big)$.

Note that in the products above we satisfy the constraint that $j \neq i$ because we know that $x_{i} = l$. The final update expression for the containment potential is then

\begin{equation}
\sum \limits_{\vec{x}_{c} | x_{i} = l}  \Big(\prod \limits_{j \in c, j \neq i} Q_{j}(x_{j} = l) + Q_{j}(x_{j} = l') \Big) *w_{\text{low}}(l) + \Big( 1 - \Big( \prod \limits_{j \in c, j \neq i} Q_{j}(x_{j} = l) + Q_{j}(x_{j} = l') \Big) \Big)* w_{\text{high}}
\end{equation}

\subsection{Attachment}
\label{sec:attachment}

\par Our approach to forming a potential that measures the presence of attachment between parts requires more care. We first consider the problem of what cliques the attachment potential $\psi_{a}$ will be measured over. Attachment is measured across boundaries, and yet finding the boundaries of objects is precisely the obstacle we are trying to overcome. In keeping with our superpixel formulation we consider pairs of superpixels, and estimate the distance between them by calculating the euclidean distance between their centroids. If this distance falls below a certain threshold $d$, then we know that the superpixels are neighbors in a loose sense. Information on how the threshold is set may be found in the appendix \ref{appen:thredhold_d}.

Thus, to determine the cliques for the attachment term we use the following algorithm:

\begin{algorithm}
\caption{Determine Attachment Cliques in Image}\label{attachment_alg}
\begin{algorithmic}[]
\For {each superpixel pair $\text{sp}_{1}, \text{sp}_{2}$}
    \If {$l_{1} \sim l_{2}$}
        \If {distance(centroid($\text{sp}_{1}$), centroid($\text{sp}_{2}$)) $> d$}
        	    \State Include clique $c = \text{sp}_{1} \cup \text{sp}_{2}$ in $\mathcal{C}$.
        \EndIf
    \EndIf
\EndFor
\end{algorithmic}
\end{algorithm}

\begin{figure}
\centering
\includegraphics[width=.2\linewidth, scale=0.3]{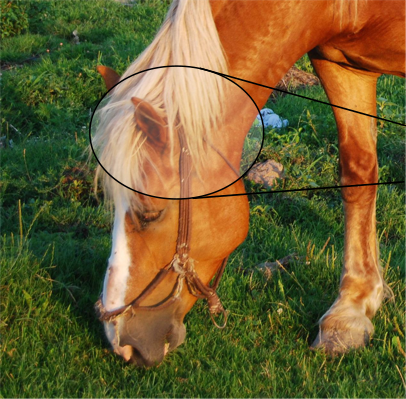}
\includegraphics[width=.35\linewidth]{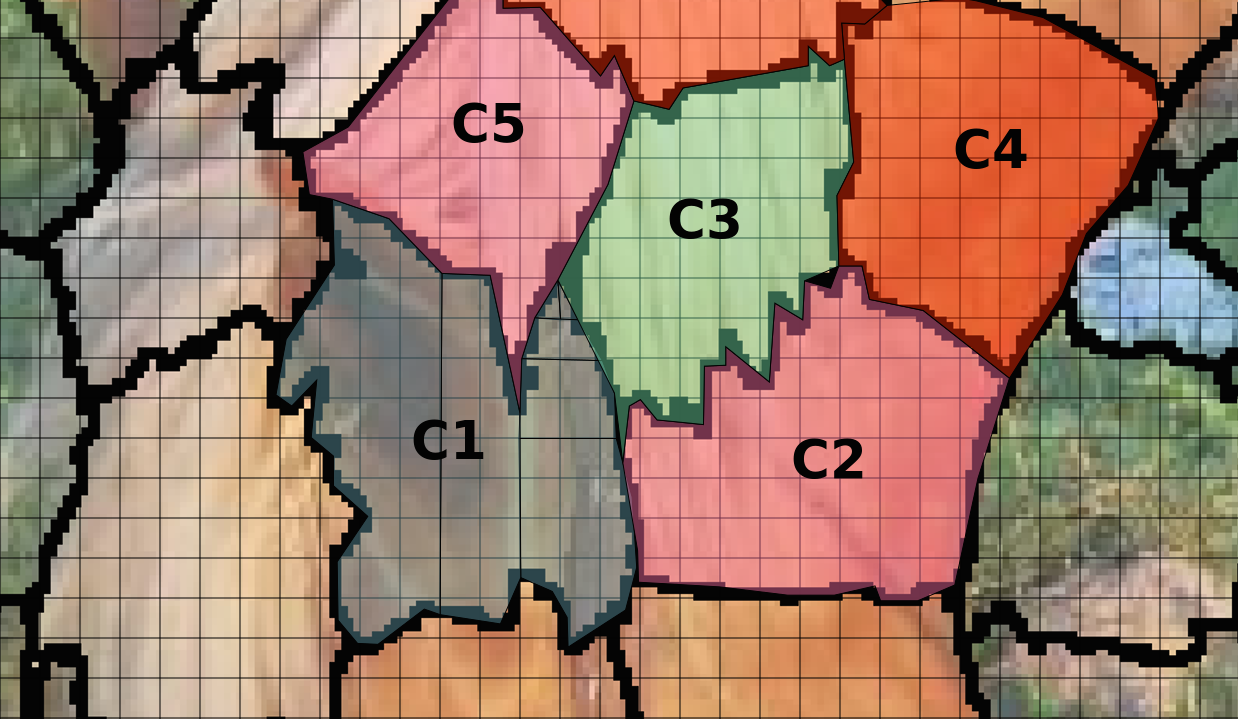}
\includegraphics[width=.35\linewidth]{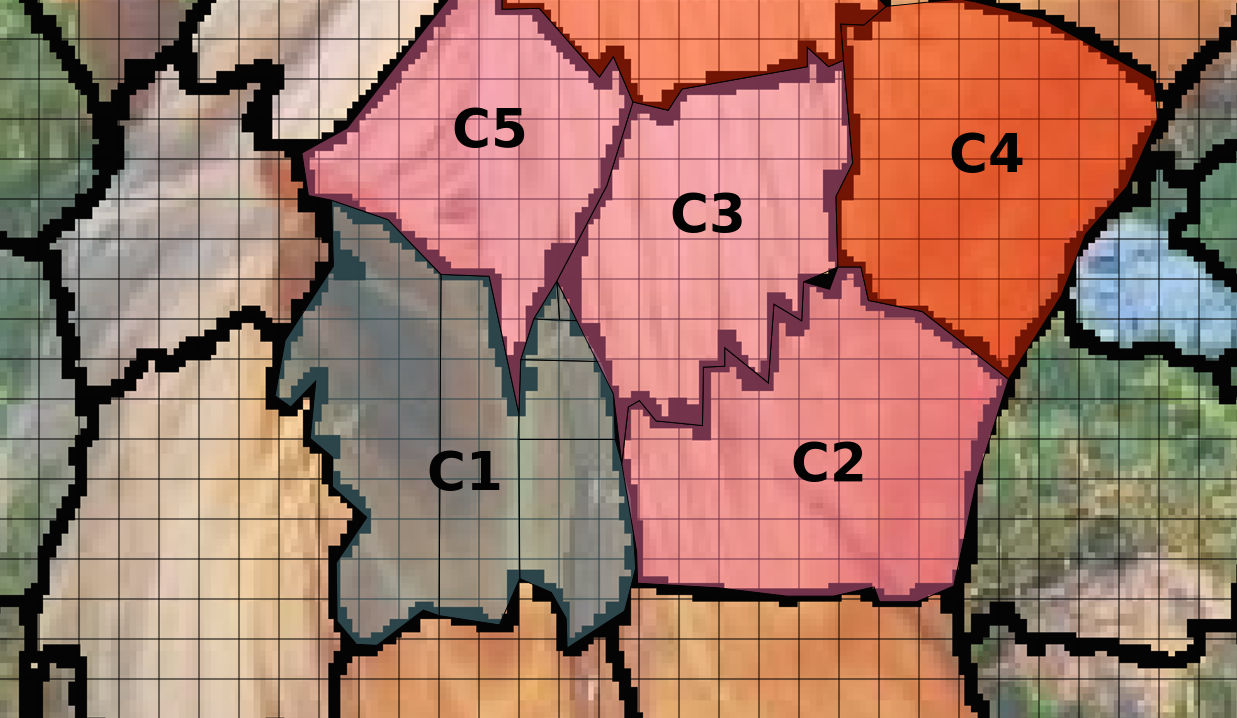}
\caption{\captionsize
       Error correction using the attachment potential. Left: The original image. Middle: A close-up of the neck, with superpixel overlay. The pink indicates a head labeling, while orange indicates neck and green indicates torso. The attachment potential corrects this error by computing the distance between the centroids (marked by c) and checking for a relation. For example, c3 is close enough to c5, c4, and c2 to check for a relation, but not to c1. Right: The corrected labeling.}
\label{fig:attachment_correction}
\end{figure}

For notational simplicity, let $c_{1} = \text{sp}_{1}$ and $c_{2} = \text{sp}_{2}$ so that $c = <c_{1},c_{2}>$. We define the attachment potential over each pair $c$ as

\begin{equation}
\psi_{\text{a}}(\vec{x}_{c}) = \begin{cases}
    w_{\text{low}}(l_{1}, l_{2}), & \forall j \in c_{1}, \vec{x}_{c_{1}} = l_{1}, \forall k \in c_{1}, \vec{x}_{c_{2}} = l_{2}\\
    w_{\text{high}}, & \text{otherwise}
  \end{cases}
\label{eq:attachment_potential}
\end{equation}

where $l_{1}$ and $l_{2}$ are the most frequent labels in the sets $\vec{x}_{c_{1}}$ and $\vec{x}_{c_{2}}$, respectively. In cases where $l_{1} \neq \text{most frequent label}(c_{1})$ or $l_{2} \neq \text{most frequent label}(c_{2})$, $w_{\text{low}}(l_{1}, l_{2}) = w_{\text{high}}$. Note that the condition for assigning a low weight is strict in the sense that we do not allow any label sharing between the two superpixels within the clique. Figure  \ref{fig:attachment_correction} illustrates how centroids are key to the formation of an attachment clique.

To compute the mean field update for the attachment term, we again begin with the update rule in Eq. \ref{eq:mean_field_update} and consider the update at pixel $x_{i}$ for label $l$. For what distributions of $Q$ would we assign a low weight? If $i \in c_{1}$ then applying the mean field assumption gives the joint distribution over the pixels in $\vec{x}_{c}$ to be $\prod \limits_{j \in c_{1}, j \neq i} Q_{j}(x_{j} = l_{1}) \prod \limits_{k \in c_{2}} Q_{k}(x_{k} = l_{2})$. Likewise, if $i \in c_{2}$ and we assign a low weight then we have $\prod \limits_{j \in c_{1}} Q_{j}(x_{j} = l_{1}) \prod \limits_{k \in c_{2}, k \neq i} Q_{k}(x_{k} = l_{2})$.

We introduce an indicator function $\mathbbm{1}_{l \in \{ l_{1}, l_{2} \}}$ to denote whether the label $l$ that we are updating over is equal to either $l_{1}$ or $l_{2}$. Putting the two expressions together gives the following distribution for assigning a low weight

\begin{equation}
\mathbbm{1}_{l = l_{1}} \Big[ \prod \limits_{j \in c_{1}, j \neq i} Q_{j}(x_{j} = l_{1}) \prod \limits_{k \in c_{2}} Q_{k}(x_{k} = l_{2}) \Big] + \mathbbm{1}_{l = l_{2}} \Big[ \prod \limits_{j \in c_{1}} Q_{j}(x_{j} = l_{1}) \prod \limits_{k \in c_{2}, k \neq i} Q_{k}(x_{k} = l_{2}) \Big]
\label{eq:gamma_exp}
\end{equation}

Let expression \ref{eq:gamma_exp} be denoted by $\gamma$. In all other cases we assign a high weight, so we subtract $\gamma$ from the sum of the joint distribution of labels over clique $c$: $1-\gamma$. This leads to the following update for each pixel.

\begin{equation}
\sum \limits_{\vec{x}_{c} | x_{i} = l} Q_{c-i}(\vec{x}_{c-i}) \psi_{\text{a}}(\vec{x}_{c}) =  \sum \limits_{\vec{x}_{c}|x_{i} = l} w_{\text{low}}(l_{1}, l_{2})*\gamma + w_{\text{high}}(l_{1}, l_{2})(1 - \gamma)
\end{equation}

\section{Experimental results}
\label{sec:experimental_results}

To test and evaluate the new potential terms, we implemented a dense CRF model in which the energy function contains the terms in Eq. \ref{eq:basic_dense_crf}, with the addition of the superpixel potential in  \ref{eq:superpixel_potential}, the containment potential in Eq. \ref{eq:inside_outside_potential} and the attachment potential in Eq. \ref{eq:attachment_potential}, namely:
\begin{equation}
E(x) = \sum_{i}{\psi_{\text{u}}(x_i)} + \sum_{i,j}{\psi_{\text{p}}(x_i,x_j)} + \sum_{\vec{x}_{c}}{\psi_{\text{s}}(\vec{x}_{c})} + \sum_{\vec{x}_{c}}{\psi_{\text{i/o}}(\vec{x}_{c})} + \sum_{\vec{x}_{c}}{\psi_{\text{a}}(\vec{x}_{c})}
\label{eq:basic_dense_crf+superpixel+novel_potentials}
\end{equation}
Our inference algorithm for the dense CRF model was based on the mean field algorithm, including the update rules derived for the superpixel potential Eq. \ref{eq:superpixel_potential} in \cite{Arnab_etal_2016_ECCV} and the update rules for Eq. \ref{eq:inside_outside_potential} and Eq.~\ref{eq:attachment_potential} as derived in Sec. \ref{sec:containment} and Sec. \ref{sec:attachment} respectively. 

For the unary potentials we used the output from the DeepLab convolutional network \cite{Chen_etal_2018_TPAMI}, which is based on ResNet101 \cite{He_etal_2016_CVPR}. Our potential terms were added to the dense CRF open source implementation packaged with the DeepLab implementation, and the mean field inference algorithm was updated accordingly. For each of the three potential terms we added to the energy function we assigned weights $w_{\text{high}}$ and $w_{\text{low}}$ that were learned in a grid search manner over a validation set.

Our tests were focused on semantic segmentation of parts, as well as tests on semantic segmentation of objects for control. For part segmentation, we used the VOC part segmentation benchmark \cite{Chen_etal_2014_CVPR}, including evaluations on both coarse and fine set of parts. 
In the fine part segmentation the object is segmented into multiple parts. There are at most 21 for the horse category and 24 for the person category, including fine parts such as eye, ear, nose, etc. In the coarse part segmentation the small parts are merged, and the number of total parts is much lower (5 for the horse and 6 for person). Our assumption is that when there are a higher number of parts, the containment and attachment potentials are more effective. For object segmentation we used the PASCAL VOC object segmentation benchmark \cite{Everingham15}. While there are other datasets available for part segmentation \cite{Luo_etal_2013_CVPR}, we found Pascal Parts to have more examples than the other part datasets.

\begin{figure}
\tabcolsep 0.03cm
\noindent\makebox[\textwidth]{%
\begin{tabular}{cccccc}

  \includegraphics[trim = 0mm 0mm 0mm 0mm,clip,height=2.0cm]{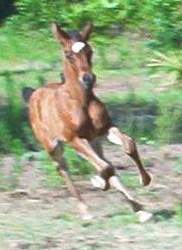} &
  \includegraphics[trim = 0mm 0mm 0mm 0mm, clip, height=2.0cm]{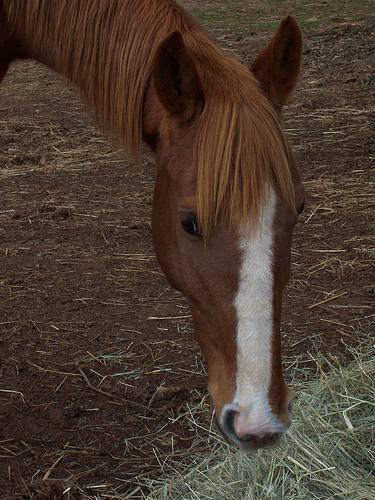} &
  \includegraphics[trim = 0mm 0mm 0mm 0mm, clip, height=2.0cm]{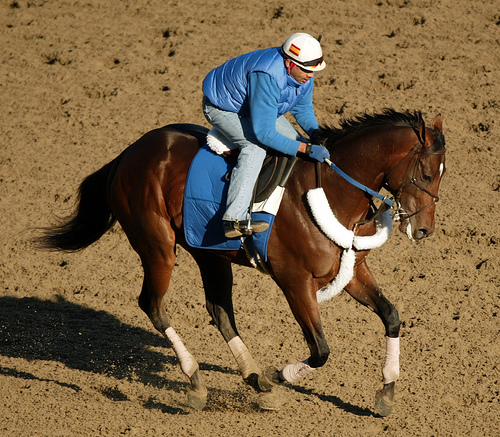} &
  \includegraphics[trim = 0mm 0mm 0mm 0mm, clip, height=2.0cm]{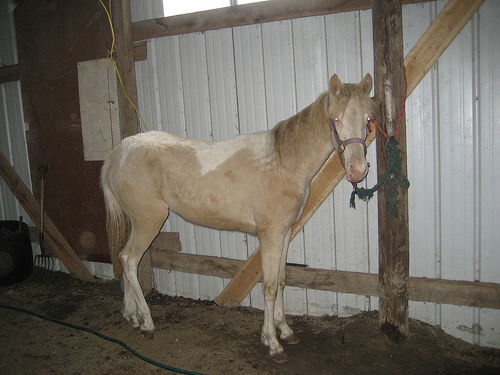} &
  \includegraphics[trim = 0mm 0mm 0mm 0mm, clip, height=2.0cm]{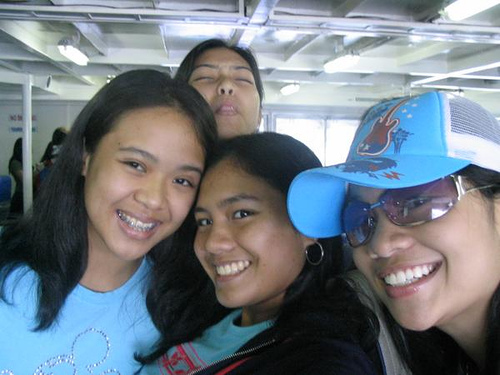} &
  \includegraphics[trim = 0mm 0mm 0mm 0mm, clip, height=2.0cm]{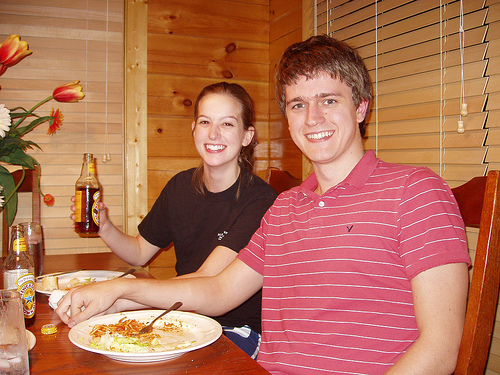} \\
  
  \includegraphics[trim = 0mm 0mm 0mm 0mm,clip,height=2.0cm]{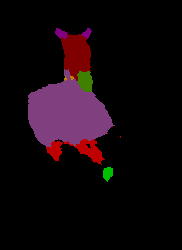} &
  \includegraphics[trim = 0mm 0mm 0mm 0mm, clip, height=2.0cm]{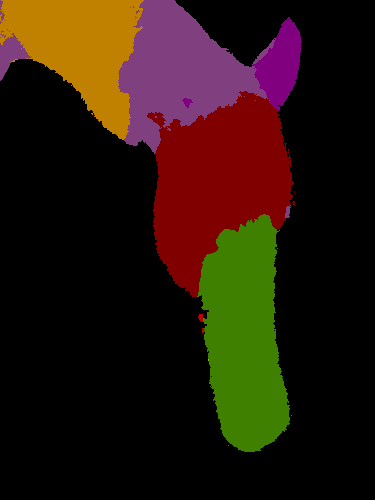} &
  \includegraphics[trim = 0mm 0mm 0mm 0mm, clip, height=2.0cm]{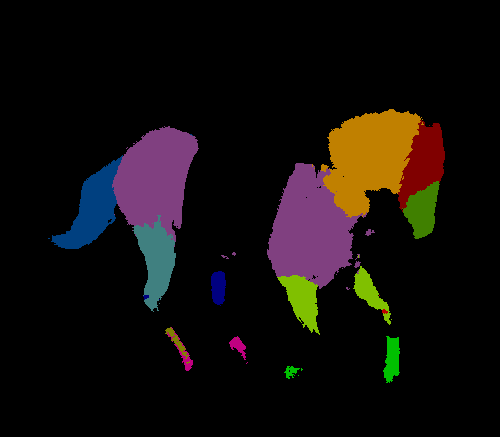} &
  \includegraphics[trim = 0mm 0mm 0mm 0mm, clip, height=2.0cm]{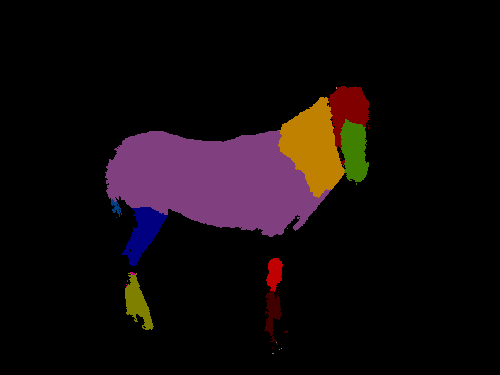} &
  \includegraphics[trim = 0mm 0mm 0mm 0mm, clip, height=2.0cm]{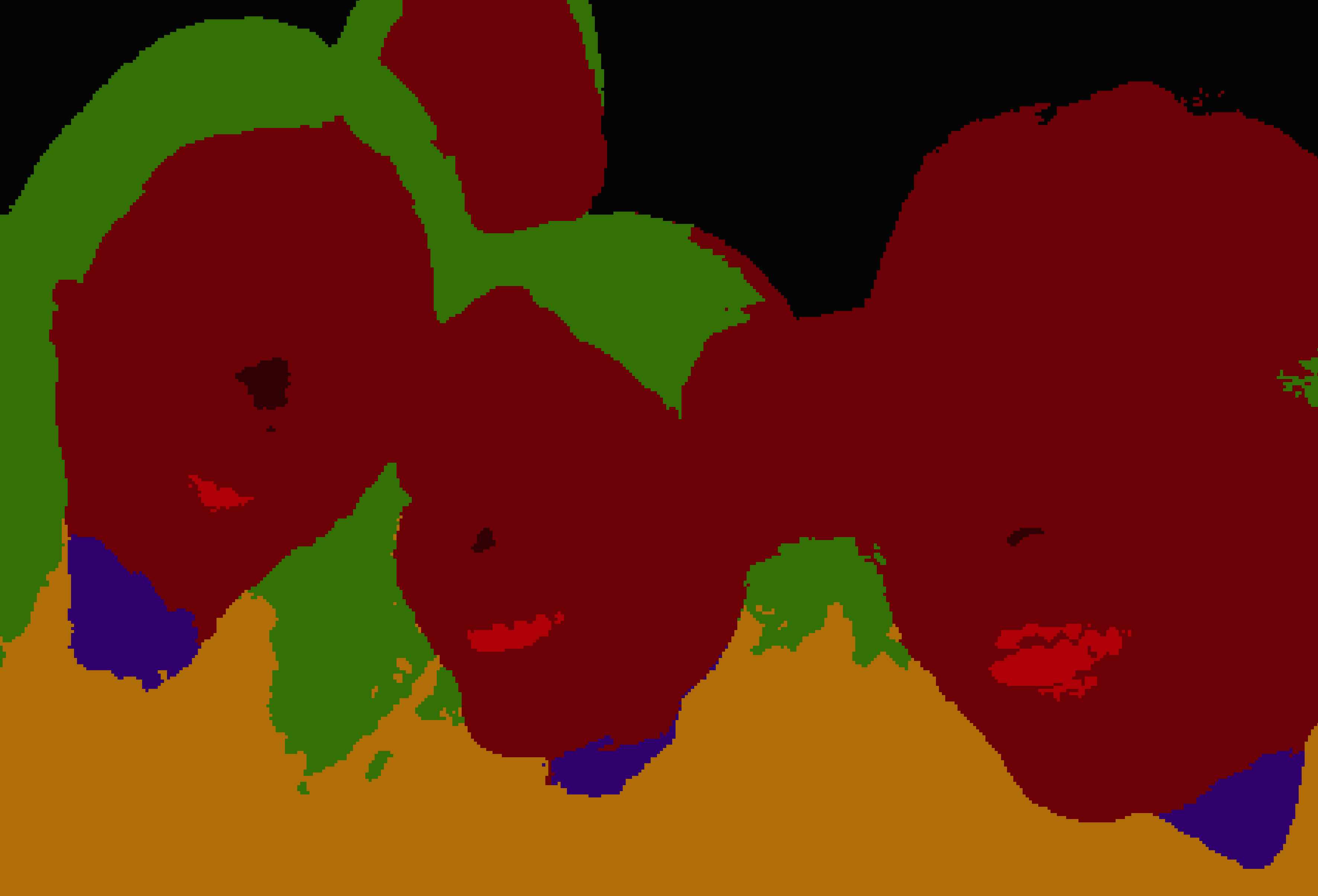} &
  \includegraphics[trim = 0mm 0mm 0mm 0mm, clip, height=2.0cm]{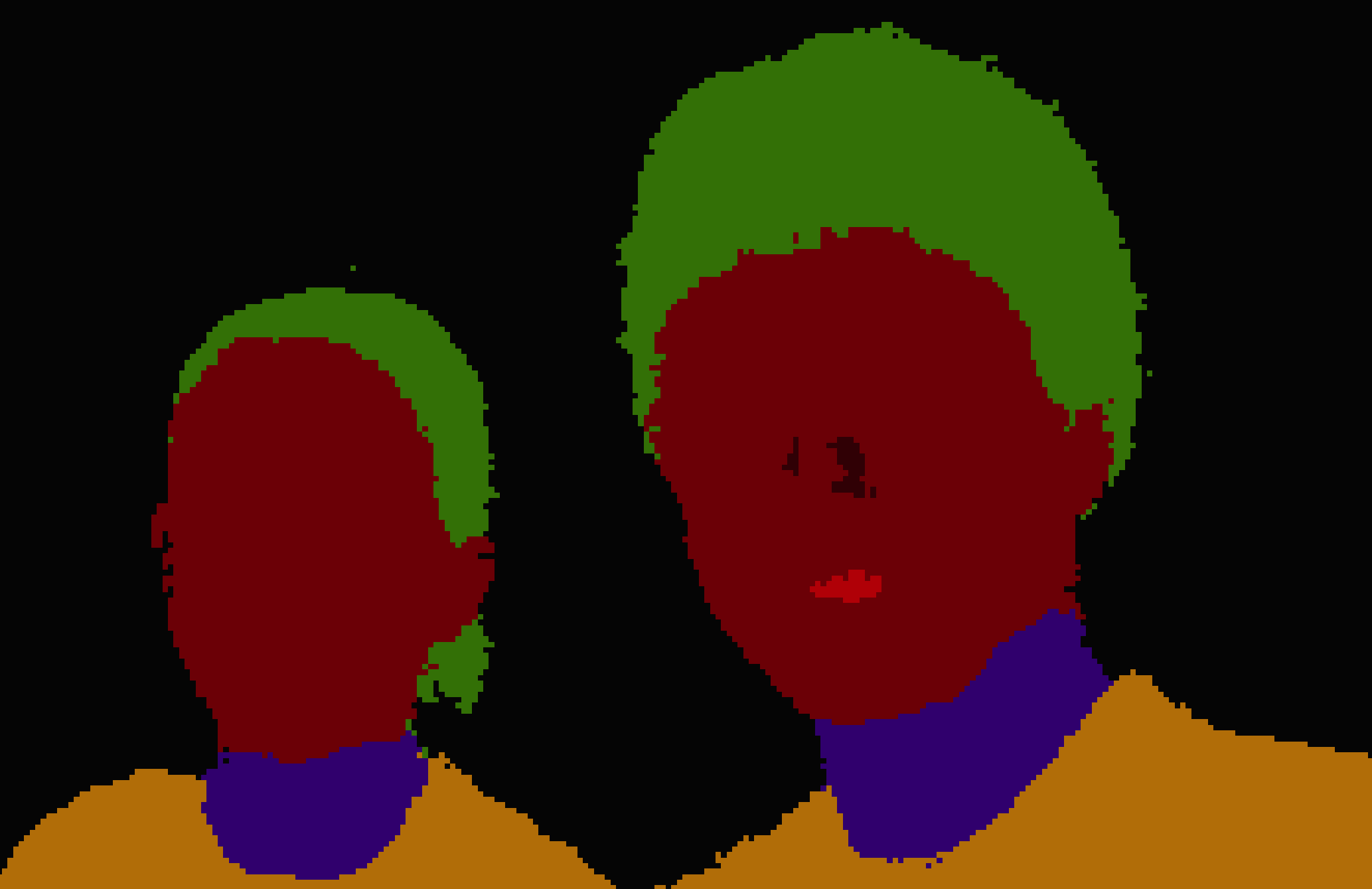} \\

  \includegraphics[trim = 0mm 0mm 0mm 0mm,clip,height=2.0cm]{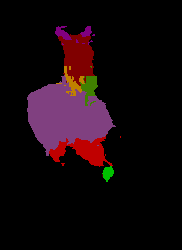} &
  \includegraphics[trim = 0mm 0mm 0mm 0mm, clip, height=2.0cm]{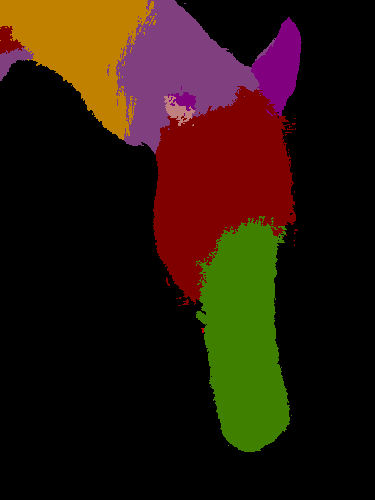} &
  \includegraphics[trim = 0mm 0mm 0mm 0mm, clip, height=2.0cm]{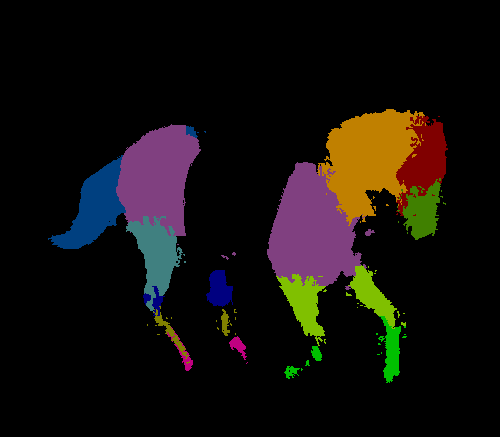} &
  \includegraphics[trim = 0mm 0mm 0mm 0mm, clip, height=2.0cm]{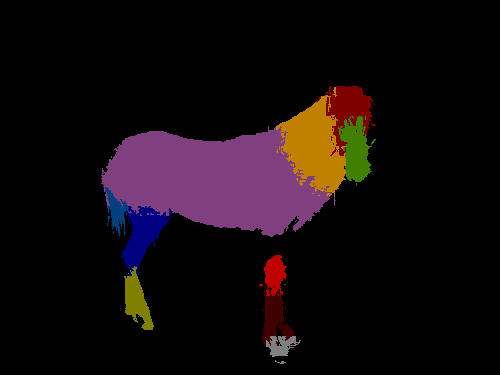} &
  \includegraphics[trim = 0mm 0mm 0mm 0mm, clip, height=2.0cm]{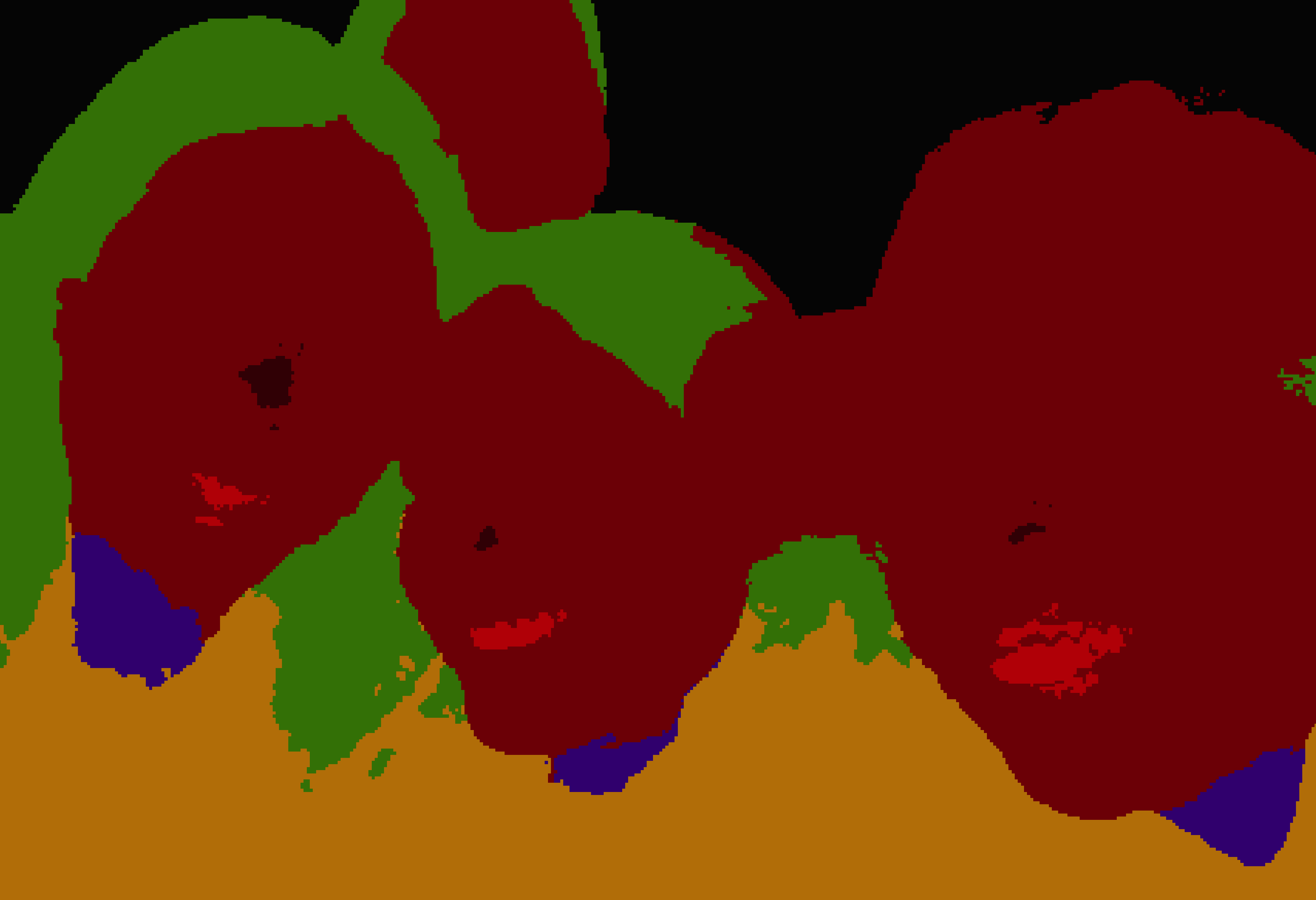} &
  \includegraphics[trim = 0mm 0mm 0mm 0mm, clip, height=2.0cm]{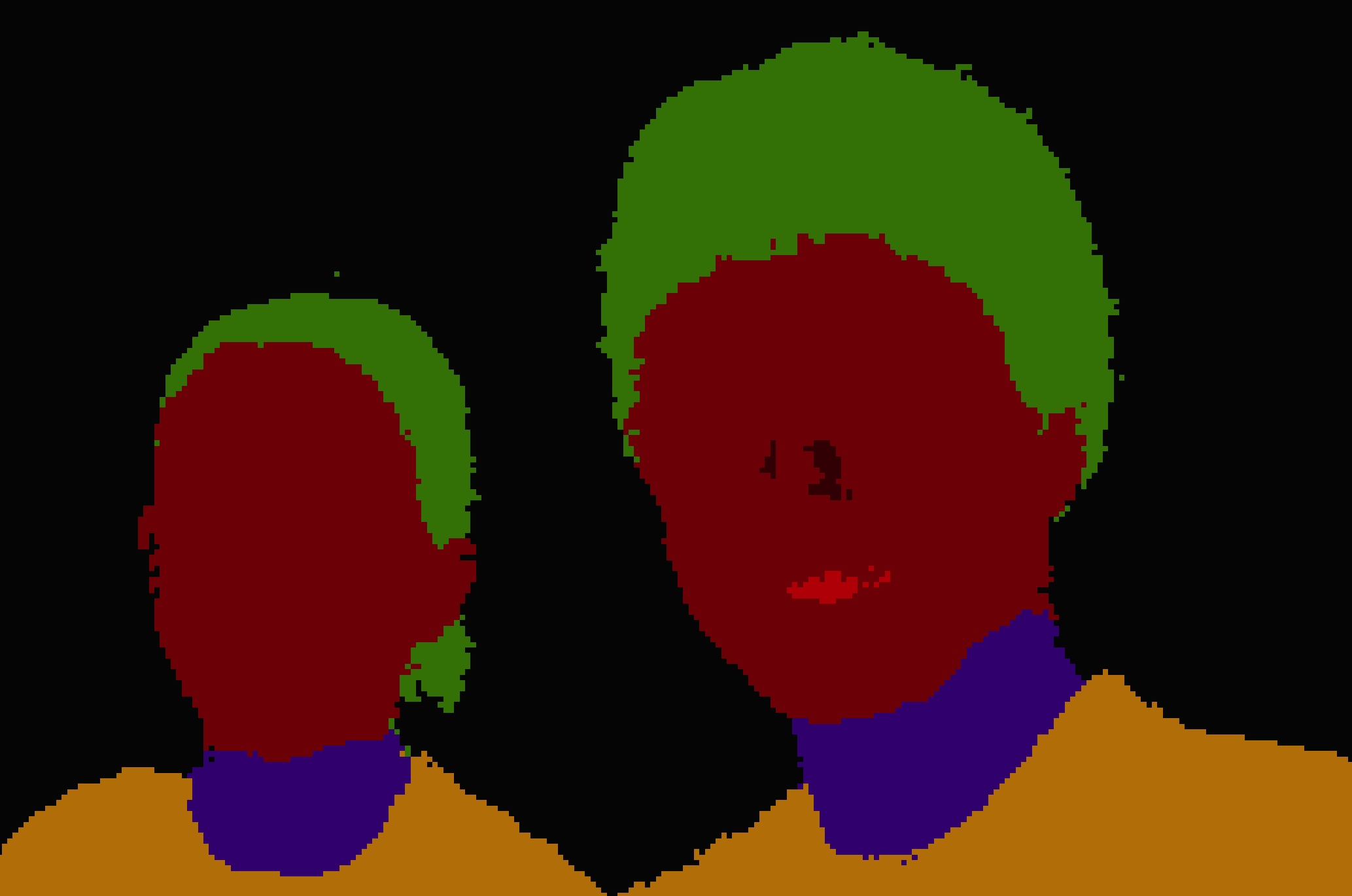} \\

A & B & C & D & E & F \\
\end{tabular}}
\caption{\captionsize
       Comparing the DeepLab ResNet101 segmentation model \cite{Chen_etal_2018_TPAMI} with (bottom row) and without (middle row) the attachment and containment terms. The original image is in the top row. \textbf{(A-D).} The attachment potential improves localization of small parts: In (A) the head is no longer touching the torso and the neck is between the two, in (B) the right ear is now partially segmented, in (C) the legs are more segmented and are accurately attached, in (D) the hoofs are now segmented. 
\textbf{(E-F).} The containment potential improves the edges of the mouth and nose in the person dataset by increasing their size.
}
\label{fig:results_attachment_highlights}
\end{figure}

Our tests on the part segmentation task included the horse category and the person category, since there are more available fine ground truth annotations in these categories than in others. For the horse category, we used the 325 images for training and performed 4-fold validation, with 100 images for testing. For the person category we used 1985 images for training with 4-fold validation and 1515 for testing. We compared our segmentation results on the test set with several state-of-the-art part segmentation algorithms, namely the DeepLab system based on ResNet101 \cite{Chen_etal_2018_TPAMI}, HZAN \cite{Xia_etal_2016_ECCV}, and RefineNet \cite{Lin_etal_2017_CVPR}. Both HAZN and RefineNet use higher-order potentials in a dense CRF framework. A comparison of our Intersection over Union (IOU) segmentation results as compared to the state-of-the-art is shown in the tables below. In Table \ref{tab:pascal_coarse_results} we note that HAZN uses only four parts for the horse, whereas we used five. Several examples from each segmentation scheme are shown in Fig. \ref{fig:results_attachment_highlights}. 

An evaluation of our model shows that the contribution of the attachment and containment potential terms is higher for the fine part segmentation task, where we achieve better results with the attachment term and comparable to the state-of-the-art results with the containment term. In the coarse part segmentation and object segmentation tasks, our model is comparable with the best current models, such as Deeplab ResNet101 \cite{Chen_etal_2018_TPAMI}. These results highlight the role of more complex high-order relations in structured prediction models for part segmentation, where the number of object parts is large.

\begin{table}
\begin{centering}
\setlength\tabcolsep{4pt}
\begin{minipage}{0.48\textwidth}
\centering
  \caption{Coarse Pascal Parts (\% IOU)}
  \begin{tabular}{lll}
    \toprule
    Model & Horse & Person \\
    \midrule
    HAZN       & 72.36  & 57.54 \\
    Graph\_LSTM & –       & 60.35 \\
    RefineNet  & –       & 68.6 \\
    DeepLab+denseCRF & 60.663 & 59.907 \\
    Superpixel & 60.536 & 59.895 \\
    Attachment & 60.682 & 57.725 \\
    \bottomrule
    \hline
  \end{tabular}
\label{tab:pascal_coarse_results}
\end{minipage}
\hfill
\begin{minipage}{0.5\textwidth}
\centering
  \caption{Fine Pascal Parts (\% IOU)}
  \begin{tabular}{lll}
    \toprule
     Model     & Horse     & Person \\
    \midrule
    DeepLab+denseCRF & 22.489 & 19.685 \\
    Superpixel & 22.466 & 19.688 \\
    Containment & 22.489 & 19.655 \\
    Attachment & 23.380 & 19.681 \\
    \bottomrule
    \hline
  \end{tabular}
 \label{tab:pascal_fine_results}
\end{minipage}
\end{centering}
\end{table}


\section{Discussion}
\label{sec:discussion}

\par In this paper we studied structural prediction models, which follow an FCN stage for the fine semantic segmentation of object parts. Our study focused on the dense CRF graphical model, which is currently the dominant graphical model coupled with FCNs for semantic segmentation. Dense CRFs are often used together with deep convolutional networks to achieve accurate detection of boundaries, and are integrated either as a separate stage after the CNNs or integrated in an end-to-end fashion. To solve the inference problem in an undirected graphical model, we have used the mean field approach, which is commonly taken in dense CRF models for segmentation.  

\par Our study included the evaluation of binary and higher-order potential terms in the CRF that are more complex than previously used for semantic segmentation. Our motivation to study such terms is derived from studies of human-like detection and localization of parts in object images. In particular we derived CRF formulations for two prominent relations from human studies, namely containment and attachment of parts. Our results show that when the segmentation task involves the localization of multiple fine-level parts, the contribution of relational features such as containment or attachment increases. This motivates further work on complex relations in a dense CRF to study the segmentation of fine parts, aimed towards replicating the richness and accuracy of human-level part segmentation. Fine part segmentation could be an avenue for solving difficult visual tasks, such as understanding object-agent and agent-agent interactions. 

A major effort in semantic segmentation research is focused on different architectures for the decoding stage in Fully Convolutional Networks.
The decoding stage usually includes mechanisms such as deconvolutions, unpooling, skip connections, pyramid pooling, gating, etc. Our work suggests that the decoding architecture in semantic segmentation networks should have the capacity to learn complex relations between pixels and superpixels, such as containment, attachment, and others. Here we proposed an implementation based on undirected graphical models, but it would be interesting to further explore network architectures to compute such relations in FCN fashion. 
Such architectures may include lateral (recurrent) or feedback connections, to compute the complex (and computationally expensive) relations selectively, given some prior. First steps in this direction were shown by \cite{Zheng_etal_2015_CVPR} for dense CRFs with the basic relations in Eq. \ref{eq:basic_dense_crf}, which were integrated in end-to-end deep Recurrent Neural Networks.

\section{Appendix: Determining threshold d}
\label{appen:thredhold_d}

The threshold $d$ by which we measure that two superpixels may satisfy attachment can be determined from the average size of the superpixels. In generating the superpixels we have imposed a regularity constraint so that each superpixel is connected and in general they have a similar shape. The threshold $d$ can then be set to the average width of each superpixel.

\subsubsection*{Acknowledgments}
This work was supported in part by the Center for Brains, Minds, and Machines, NSF STC award 1231216, as well as the MIT-IBM Brain-Inspired Multimedia Comprehension project. 

\bibliographystyle{plain} 


\end{document}